\documentclass[a4paper]{llncs}
\usepackage{graphicx}
\renewcommand{\thetable}{\arabic{table}}
\usepackage{color}
\usepackage{amsmath}
\usepackage{amssymb}
\usepackage{graphicx}
\usepackage{textcomp}
\usepackage{multirow}
\usepackage{xcolor}
\usepackage[numbers]{natbib}
\usepackage[bookmarks,hidelinks]{hyperref}
\providecommand{\tabularnewline}{\\}
\newcommand\correspondingauthor{\thanks{Corresponding author.}}
\usepackage{geometry}
\begin{document}
\nocite{*}
%
% \title{An extended retrieval method for improving LLM's long-context question-answering and summarization tasks\thanks{Supported by Phenikaa University.}}
\title{BERT-VBD: Vietnamese Multi-Document Summarization Framework}
%
% \titlerunning{Proceedings of the 13th International Conference on Information Technology and Its Applications (CITA 2024)}
% If the paper title is too long for the running head, you can set
% an abbreviated paper title here
%
\author{Tuan-Cuong Vuong\inst{1} %\orcidID{0000-0002-0999-1950} 
\and
Trang Mai Xuan\inst{1}%\orcidID{1111-2222-3333-4444} 
\and
Thien Van Luong\inst{1}\correspondingauthor%\orcidID{2222--3333-4444-5555}
}
\authorrunning{Cuong et al.}
% First names are abbreviated in the running head.
% If there are more than two authors, 'et al.' is used.
%
\institute{Phenikaa University, Ha Noi 100000, Viet Nam
\email{\\21011490@st.phenikaa-uni.edu.vn \\
\{trang.maixuan,thien.luongvan\}@phenikaa-uni.edu.vn}}
\maketitle              % typeset the header of the contribution

\begin{abstract}
In tackling the challenge of Multi-Document Summarization (MDS), numerous methods have been proposed, spanning both extractive and abstractive summarization techniques. However, each approach has its own limitations, making it less effective to rely solely on either one. An emerging and promising strategy involves a synergistic fusion of extractive and abstractive summarization methods. Despite the plethora of studies in this domain, research on the combined methodology remains scarce, particularly in the context of Vietnamese language processing. This paper presents a novel Vietnamese MDS framework leveraging a two-component pipeline architecture that integrates extractive and abstractive techniques. The first component employs an extractive approach to identify key sentences within each document. This is achieved by a modification of the pre-trained BERT network, which derives semantically meaningful phrase embeddings using siamese and triplet network structures. The second component utilizes the VBD-LLaMA2-7B-50b model for abstractive summarization, ultimately generating the final summary document. Our proposed framework demonstrates a positive performance, attaining ROUGE-2 scores of $39.6\%$ on the VN-MDS dataset and outperforming the state-of-the-art baselines.
% This paper introduces an innovative MDS framework comprising two interconnected components within a pipeline architecture, seamlessly integrating both extractive and abstractive approaches tailored for the Vietnamese language. 

\end{abstract}

\section{Introduction}

Multi-Document Summarization (MDS) aims to distill information from multiple documents into a concise representation, preserving key points and eliminating redundancy. In the Vietnamese language context, MDS presents a unique challenge due to its nascent stage and inherent linguistic complexities. Traditionally, topics are explored through diverse Vietnamese articles, each offering distinct perspectives. While directly extracting excerpts can facilitate initial understanding, it often leads to incoherence. Conversely, purely abstractive approaches might struggle to retain salient details. Combining extractive and abstractive approaches holds significant potential for Vietnamese MDS problems. However, research on this method is scarce, both in English and Vietnamese. Existing studies, like those by Y Gao et al.~\cite{gao2021extractive} and H Jin et al.~\cite{jin2020multi}, perform extractive and abstractive summarization in parallel with separate input representations for each model, rather than truly combining them with a shared input.
% The combining extractive and abstractive approach is very potential for the Vietnamese MDS problem. However, research on this approach is still very limited, whether it’s English or Vietnamese. Some studies have performed extractive and abstractive summarization in parallel with different input representations for each model (such as Y Gao et al~\cite{gao2021extractive} and H Jin et al~\cite{jin2020multi}), rather than combining extractive and abstractive approach and use the same input.

Our paper proposes a Vietnamese MDS framework utilizing a two-component pipeline architecture that merges extractive and abstractive approaches. This combination offers three key benefits: firstly, it preserves crucial information for the reader using the extractive method, while simultaneously presenting it in a more reader-friendly, concise form through the abstractive approach. Secondly, it optimizes resource usage compared to solely relying on abstractive summarization, which can be computationally expensive when retaining important details. Finally, leveraging the pre-trained models available in the extractive approach enhances the summarization ability by incorporating more information from both sentence and word embeddings.
% In this paper, we propose a Vietnamese MDS frame which contains two components in a pipeline architecture combining extractive and abstractive approach. This combination has three main benefits: first, it retains important information by extrac- tive approach for the reader and at the same time summarizes it into a text that is close to the reader by abstractive approach; second, it can optimize resources because if you only use the abstractive approach for the summarization problem, it can consume a lot of resources to retain important information for the reader; third, the combination of training models available in the extractive approach helps to increase the summarization ability because there is more information from embedding sentences and embedding words.

In summary, our key contributions are twofold:

\begin{itemize}
    \item We propose a novel framework that merges extractive and abstractive summarization techniques to tackle Vietnamese MDS challenges. This framework leverages the deep learning models such as Sentence-BERT (SBERT) \cite{reimers2019sentence} and VBD-LLaMA2-7B-50b~\cite{VBDLLaMA2Chat2024} to generate a final summary that both retains important content and ensures readability.
    % We propose a new frame that combines Extractive and Abstractive summarization to summarize Vietnamese Multi-document by applying deep learning model Sentence-BERT (SBERT)~\cite{reimers2019sentence} and VBD-LLaMA2-7B-50b model to create a final summary that retains important content and is easy to read.
    \item We evaluate the proposed framework on the VN-MDS\footnote{https://github.com/lupanh/VietnameseMDS} dataset. Our experiments demonstrate promising results, showcasing its originality and correctness by effectively combining various methods. We conduct a comprehensive comparison between our framework and the current models applied to the Vietnamese language, which shows that our model is superior to them.
\end{itemize}
\section{Related Work}
In the domain of document summarization, two dominant approaches exist: extractive and abstractive. Extractive methods identify and extract key text segments like words, phrases, or sentences to form the summary. On the other hand, abstractive methods generate entirely new text summaries encapsulating the core information from the original documents. A recently emerged hybrid approach integrates both extractive and abstractive methods, aiming to produce summaries of even higher quality. This hybrid approach typically involves extracting a subset of crucial sentences using the extractive method, followed by the abstractive method's application on these selected sentences to generate the final summary (Liu et al.~\cite{liu2021combined}).
% Generally, there are two approaches to document summarization: extractive approaches, where salient text segments like words, phrases or sentences are identified and extracted to form the summary, and abstractive approaches, which generate new summary text containing key information from the source documents. There is also a hybrid approach combining extractive and abstractive methods, an emerging technique. Integrating extractive and abstractive can produce higher quality summaries. Broadly, the extractive method will extract a certain number of salient sentences, then the abstractive method is applied on the extracted sentences to generate the final summary \cite{liu2021combined}.

Extractive summarization involves extracting salient sentences or phrases from documents and concatenating them into a summary. The current approaches often employ graph-based techniques such as LexRank~\cite{erkan2004lexrank} and TextRank~\cite{mihalcea2004textrank} using sentence embeddings or focus on features such as sentence position and term frequency to calculate importance. Recent methods apply machine learning models such as reinforcement learning~\cite{narayan2018ranking} and deep learning~\cite{verma2017extractive} to identify key phrases for summarization. Devlin et al~\cite{devlin2018bert} proposed the BERT model and Liu et al~\cite{liu2019roberta} proposed the RoBERTa model, two models that have achieved state-of-the-art performance on sentence-pair regression tasks like semantic textual similarity. Reimers et al.~\cite{reimers2019sentence} proposed SBERT which enables encoding sentences and paragraphs into dense vector representations using pretrained language models. It achieves state-of-the-art results on various sentence embedding benchmarks. In this work, we utilize SBERT for the extractive summarization process.

Unlike extractive methods that directly copy key segments from the source text, abstractive summarization leverages deep learning capabilities in natural language processing (NLP) to generate entirely new sentences or paraphrases for the summary, even using words not present in the original documents. This recent advancement is largely driven by the increasing complexity and power of language models. Early attempts such as Nallapati et al~\cite{nallapati2016sequence} employed recurrent neural networks (RNNs) with attention mechanisms, incorporating auxiliary techniques such as keyword mapping and rare word extraction. Building upon this, Song et al~\cite{song2019abstractive} proposed a model utilizing long short-term memory units (LSTMs) and convolutional neural networks (CNNs) to select and combine phrases into new sentences. Sequence-to-sequence models have proven particularly successful for summarizing long texts, evidenced by BART (Savery~\cite{savery2020question}) with its bi-directional encoding and auto-regressive decoding. More recently, large pre-trained models like LLaMA~\cite{touvron2023llama1} and LLaMA-2~\cite{touvron2023llama2} from Meta, which are both auto-regressive transformer-based architectures, have further pushed the boundaries of abstractive summarization. In essence, deep learning techniques like RNNs, CNNs, attention mechanisms, and large pre-trained models have propelled abstractive summarization forward by allowing the generation of novel and informative summaries without relying solely on the source text.

% In contrast to extractive methods, abstractive summarization generates new sentences or paraphrases to produce summaries using words not in the source documents. This approach has thrived recently with deep learning for NLP due to language complexity. Nallapati et al \cite{nallapati2016sequence} propose using recurrent neural network (RNN) with attention to encode-decode with auxiliary models like keyword mapping and rare word extraction. Song et al \cite{song2019abstractive} propose an long and short memory units term (LSTM) and convolutional neural network (CNN) model to generate new sentences by selecting and combining phrases. Many researchers have applied sequence-to-sequence models successfully for long-to-short text summarization. BART \cite{savery2020question} combines bi-directional encoding and auto-regressive decoding. LLaMA \cite{touvron2023llama1} and LLaMA-2 \cite{touvron2023llama2} from Meta are auto-regressive transformer-based language models for abstractive summarization. Overall, deep learning techniques like RNNs, CNNs, attention, and large pretrained models have advanced abstractive summarization by generating novel summarized content.

There has been limited research combining extractive and abstractive summarization models for Vietnamese. Liu et al.~\cite{liu2021combined} proposed a two-stage method utilizing both approaches. First, they extract important sentences using sentence similarity matrices or pseudo-titles, considering features such as  position and structure. This identifies coarse-grained salient sentences. Second, they abstractively restructure and rewrite the extracts using beam search to generate new summary sentences. These then act as pseudo-summaries for the next round, with the final pseudo-title as the summary. Experiments show improved results over either approach alone.
Tretyak and Stepanov~\cite{tretyak2020combination} also presented a method for long document summarization using both techniques. They first extractively select content using pre-trained transformer language models to condition the abstractor. The abstractor then rewrites the summary abstractively. Jointly applying both approaches significantly improves summarization quality and ROUGE scores compared to using either in isolation.

In Vietnam, most summarization research has focused on single-document extractive methods. Quoc To et al.~\cite{to2021monolingual} concatenated documents into paragraphs and passed sentences into BERT for clustering, ranking and extraction. Nguyen et al.~\cite{nguyen2018towards} evaluated unsupervised, supervised and deep learning techniques on the VN-MDS and ViMs datasets, but did not propose any models for performance improvements. Manh et al.~\cite{manh2019extractive} proposed using K-means clustering combined with centroid-based, MMR and position-based methods. They pre-processed input into sentence vectors and extracted summaries by selecting informative sentences. Thanh et al.~\cite{thanh2022graph} proposed combining graph methods and PhoBERT to generate readable summaries retaining key content, but not compared with models that combine abstractive and extracive methods.

\section{Proposed Method}

In this section, we delineate the proposed model architecture and approach. In particular, in Section \ref{subsec:overview}, we provide a high-level schematic of the proposed end-to-end model pipeline. Next, we detail the data pre-processing steps taken to prepare the input data for summarization in Section \ref{subsec:datapreprocess}. In Section \ref{subsec:extractive-method}, we describe the extractive summarization techniques leveraged to identify and extract salient content from the source documents. Subsequently, we explain the abstractive summarization components, which condense and paraphrase the extracted content to generate novel phrasing. Finally, we discuss the post-processing phase where the extracted and generated portions are combined and refined to produce the final summarized output in Section \ref{subsec:abtractive-method}.

% In this section, we will cover the content and how the model we propose works. In part 1, we present an overview of the operation diagram of the model. In part 2, we present the data pre-processing process. In part 3, we present the extractive summarization algorithm. Finally, we will cover the abstractive summarization model and how postprocessing to give the final result in Section 4.

\subsection{Overview}\label{subsec:overview}

\begin{figure}[ht]
    \centering
    \includegraphics[width=1\columnwidth]{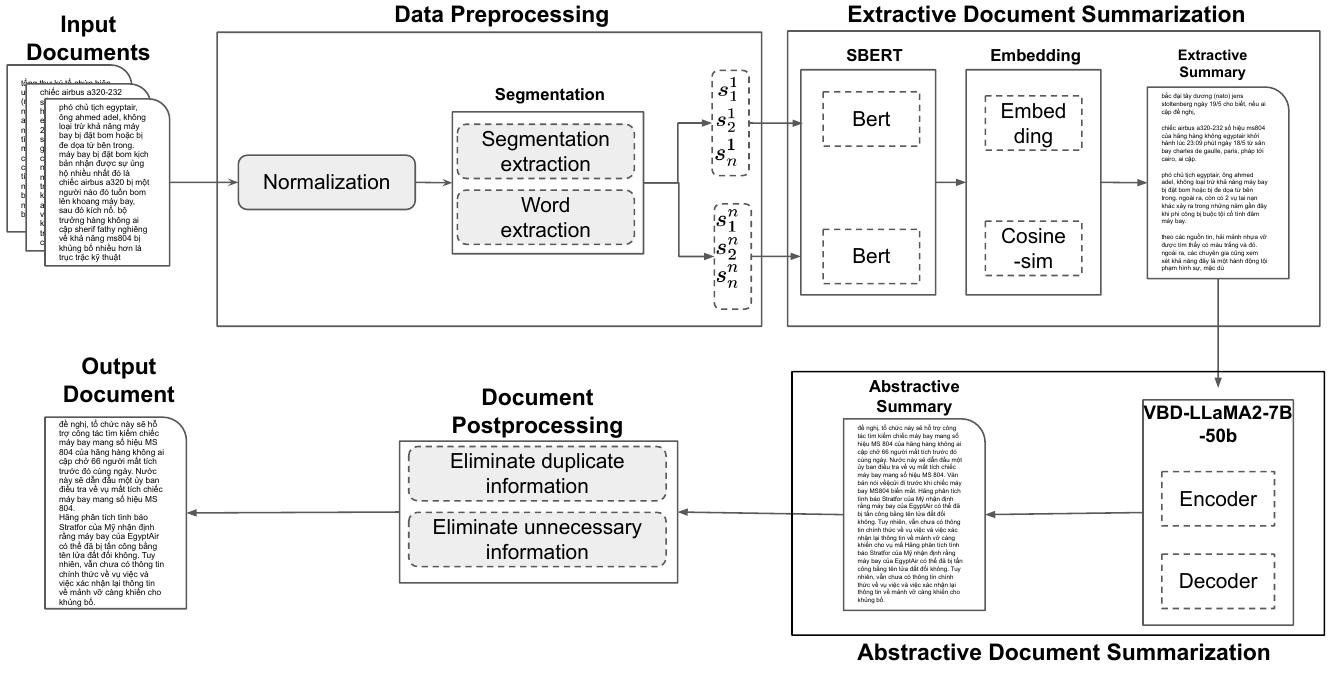}
    \caption{Pipeline for proposed model.}
    \label{fig:pipeline}
\end{figure}

We propose a novel MDS model for Vietnamese that combines extractive and abstractive approaches in a pipeline architecture. The model contains two components: extractive summarization followed by abstractive summarization. The abstractive component takes the output of the extractive component as input. This hybrid architecture aims to leverage the advantages of both approaches while overcoming their limitations. As illustrated in Fig.~\ref{fig:pipeline}, the extractive component identifies and extracts salient content from the source documents. The abstractive component then condenses and rewrites this extracted content to generate a novel summarized text. The pipeline design allows our model to utilize the strengths of extractive selection and abstractive rewriting for Vietnamese MDS. Details of each component are presented in the following sections.

% We propose a new MDS model which contains two com- ponents combining extractive and abstractive approaches for Vietnamese MDS. These two components are built according to the pipeline architecture, abstractive summarization component takes the output of the extractive summarization component as its input. This hybrid architecture allows the model to take advantages and overcome the disadvantages of both approaches. Our model is illustrated in Fig~\ref{fig:pipeline}, details of the components are presented in the following sections.

\subsection{Data Pre-processing}\label{subsec:datapreprocess} 

Data pre-processing is critical as it directly impacts model efficiency and performance. Due to the complex properties of Vietnamese, specialized pre-processing is required including cleaning, normalization, and enrichment.

\textbf{Data Normalization: }Cleaning involves lowercase conversion and removing meaningless words and non-alphanumeric characters. Normalization creates uniformity by involving the following steps:

\begin{itemize}
    \item Convert to lowercase: For cleaning and normalization, words are converted to lowercase to create uniformity between upper and lower case versions with the same meaning.
    \item Eliminate words that do not have much meaning: Meaningless words are eliminated, as the VN-MDS journalistic dataset contain mainly salient words. Therefore, non-semantic symbols such as $"\%", ";", ":"$ will be removed to save space and processing time.
    \item Non-alphanumeric characters without semantic content are removed, as they usually do not contribute meaningful information for general NLP tasks.
\end{itemize}

\textbf{Segmentation: }Segmentation is then performed including sentence splitting to divide text into sentences and word splitting to break sentences into word and phrase tokens:

\begin{itemize}
    \item Sentence splitting: This aims to divide a paragraph or document into individual sentences. It primarily relies on punctuation cues such as periods, question marks, and exclamation points to delineate sentence boundaries.
    \item Word splitting: This process breaks down larger text units such as phrases, sentences, or even entire documents into smaller pieces known as tokens. These tokens can be individual words or multi-word phrases.
\end{itemize}

\subsection{Extractive Summarization}\label{subsec:extractive-method}

In order not to lose information in each document, we will divide the documents into sentences corresponding to that document, then proceed to select the important content and classify it in clusters with the best results. Suppose, after the segmentation of the $n$ document, we will obtain the $s_n^m$ sentence, with the index $n$ corresponding to the index of the document in the input, which $m$ corresponds to the sentence index in the corresponding document. SBERT \cite{reimers2019sentence} framework allows us to convert $s_n^m$ sentences to to dense vector representations $u_n^m$ using pre-trained language models. Then we proceed to calculate sentence similarity:

\begin{equation}
    \mathrm{Sim}(s_{m}^n, s_{m+1}^n) = \frac{(u_{m}^n \cdot u_{m+1}^n)}{(||u_{m}^n|| \times  ||u_{m+1}^n||)}
\end{equation}
where $(s_{m}^n, s_{m+1}^n)$ are the two sentences to be calculated; $\mathrm{Sim}(s_{m}^n, s_{m+1}^n)$ the cosine similarity of the two sentences; $u_{m}^n$ and $u_{m+1}^n$ are vector representations for $s_{m}^n$ and $s_{m+1}^n$, respectively; $(u_{m}^n \cdot u_{m+1}^n)$ is the dot product of vectors, it measures the sum of the products of corresponding elements across both vectors; $||u_{m}^n||$ and $||u_{m+1}^n||$ are the magnitudes (lengths) of vectors $u_{m}^n$ and $u_{m+1}^n$, respectively.

After calculating the similarity of the sentences in the document, we proceeded to use the k-means algorithm combined with the elbow optimization algorithm to find the optimal number of clusters. For this, we use the $\alpha$ parameter to be able to adjust the number of input properties for the clustering model to give the final text cluster, as mentioned in Fig.~\ref{fig:pipeline}, in which, the number of input attributes corresponds to the number of embedded sentences.

\subsection{Abstractive Summarization}\label{subsec:abtractive-method}

Abstractive summarization creates the summary text as a sequence of words based on the input document sequences. We employ a sequence-to-sequence model combining transformer encoders and decoders. The encoder transforms the input sequence into a vector representation, and the decoder converts this vector into a target sequence. We evaluated three models, namely, VBD-LLaMA2-7B-50b, PhoGPT and Vistral-7B-Chat, where we found that VBD-LLaMA2-7B-50b performs the best. The encoder maps the input document extracted to a latent feature vector representation, and the decoder autoregressively generates the output summary one word at a time based on this representation. This allows producing novel phrasing and paraphrasing of the input content. Such post-processing improves the summarization performance by eliminating repetition, which can reduce coherence as VBD-LLaMA2-7B-50b may generate consecutive identical words.
\section{Performance Evaluation and Comparison}

\subsection{Hardware Configuration}\label{subsec:metrics}

The Vietnamese multi-document summarization model was built and tested on our server, whose hardware configuration is detailed in Table~\ref{tab:hardware}

\begin{table}[ht]
    \centering
    \caption{Hardware characteristics}   
    \begin{tabular}{r|l}
    \hline 
    \textbf{Unit} & \textbf{Description}\tabularnewline
    \hline 
    Processor & Intel(R) Xeon(R)  \\
              & W-2123 CPU @ 3.60GHz\tabularnewline
    \hline 
    RAM & Quadro RTX 8000 - 48GB \tabularnewline
    \hline      
    Operating System & Ubuntu 20.04.6 LTS\tabularnewline
    \hline 
    \end{tabular}
     
    \label{tab:hardware}
     
\end{table}

\subsection{Evaluation Procedure}

\subsubsection{Experiment on extractive summarization:} In this experiment, the model runs on VN-MDS data. The VN-MDS data provides a citation-based reference summary. In this process, we will refine the parameter $\alpha$, as mentioned in Section~\ref{subsec:extractive-method}. Adjusting the $\alpha$ parameter will affect the result of our model because the output of the extrative document summarization is the input of the abstractive summary module.

\subsubsection{Comparison with existing hybrid approaches:} To demonstrate the effectiveness of our hybrid extractive-abstractive summarization approach, we conduct a comparative evaluation against the state-of-the-art MDS model proposed by Thanh et al.~\cite{thanh2022graph}. This model also employs a combined extractive-abstractive strategy, making it a suitable benchmark for assessing our model's performance.

\subsubsection{Comparison with existing non-hybrid approaches:} To further isolate the contribution of our hybrid approach, we compare our model to other non-hybrid techniques that rely solely either on extraction methods or abstraction methods. For this comparison, we select MART, KL, and LSA as baseline models. These models have been recently identified as the top performers proposed in \cite{nguyen2018towards}. 

\subsection{Experiment Results}

\subsubsection{Experiment on extractive summarization:} As shown in Table~\ref{tab:extractive_results}, our experiments on the extractive document summarization phase reveal that incorporating sentence relationships based on representation vectors created by pre-trained language model improves the model's ROUGE-2 score. However, this approach negatively impacts ROUGE-1 and ROUGE-L scores compared to a graph model using only sentence-to-sentence relationships based on word frequency vectors. This suggests that using only word frequency at the morphological level for sentence representation leads to better ROUGE-1 and ROUGE-L scores. The optimal ROUGE-2 score was achieved with an $\alpha$ value of 0.2, as illustrated in Table~\ref{tab:extractive_results}, which we adopted for the extraction phase.  

\begin{table}[ht]
    \centering
    \caption{Multi-document extractive summarization model results.}
    % Preview source code for paragraph 0

    % Preview source code for paragraph 0

    {\begin{tabular}{|c|c|c|c|c|c|c|c|c|c|}
    \hline 
    \multirow{2}{*}{$\alpha$} & \multicolumn{3}{c|}{ROUGE-1} & \multicolumn{3}{c|}{ROUGE-2} & \multicolumn{3}{c|}{ROUGE-L}\tabularnewline
    \cline{2-10} \cline{3-10} \cline{4-10} \cline{5-10} \cline{6-10} \cline{7-10} \cline{8-10} \cline{9-10} \cline{10-10} 
     & Precision & Recall & F1 score & Precision & Recall & F1 score & Precision & Recall & F1 score\tabularnewline
    \hline 
    0.1 & 49.3 & 90.4 & 61.2 & 33.1 & 63.2 & 41.6 & 28.1 & 54.6 & 35.5\tabularnewline
    \hline 
    0.2 & \textbf{49.5} & 90.4 & \textbf{61.4} & \textbf{33.2} & 63.3 & \textbf{41.7} & \textbf{28.2} & 54.7 & 35.6\tabularnewline
    \hline 
    0.3 & 49.4 & 90.5 & 61.3 & 33.1 & 63.2 & 41.6 & 28.1 & 54.4 & 35.5\tabularnewline
    \hline 
    0.4 & 46.6 & 91.4 & 58.9 & 32.1 & 65.6 & 41.1 & 27.1 & 56.2 & 34.9\tabularnewline
    \hline 
    0.5 & 38.9 & \textbf{93.9} & 52.4 & 28.4 & \textbf{71.4} & 38.7 & 23.7 &\textbf{ 60.8} & \textbf{38.7}\tabularnewline
    \hline 
    \end{tabular}}

    \label{tab:extractive_results}
\end{table}

\subsubsection{Comparison with existing hybrid approaches:} To better show the effectiveness of our model, we compare with the modern model used for MDS by Thanh et al.~\cite{thanh2022graph}. Our model's comparative results show a ROUGE-1 Precision of 62.8\% compared to 61.77\% of Thanh et al.'s model. Although our Recall score is slightly lower at 79.7 compared to 79.96 of the baseline, our F1-ROUGE-1 score is significantly better at 70.1\% compared to 68.63\% of the baseline. Additionally, our ROUGE-2 scores are remarkably higher that of Thanh et al.'s model. In terms of ROUGE-L, our Precision score is lower at 28.1\% compared to 29.3\% of the baseline, however, our Recall and F1 scores are again significantly better. Overall, our model demonstrates superiority over Thanh et al.'s model based on the VN-MDS dataset.

\begin{table*}[ht]
    \centering
    \caption{Comparative results of models on VN-MDS dataset.}
    % Preview source code for paragraph 0

    % Preview source code for paragraph 0

    \resizebox{1\textwidth}{!}{\begin{tabular}{|c|c|c|c|c|c|c|c|c|c|}
    \hline 
    \multirow{2}{*}{Model} & \multicolumn{3}{c|}{ROUGE-1} & \multicolumn{3}{c|}{ROUGE-2} & \multicolumn{3}{c|}{ROUGE-L}\tabularnewline
    \cline{2-10} \cline{3-10} \cline{4-10} \cline{5-10} \cline{6-10} \cline{7-10} \cline{8-10} \cline{9-10} \cline{10-10} 
     & Precision & Recall & F1 score & Precision & Recall & F1 score & Precision & Recall & F1 score\tabularnewline
    \hline 
    Thanh et al.~\cite{thanh2022graph} & 61.77 & \textbf{79.96} & 68.63 & 31.36 & 40.7 & 34.89 & \textbf{29.3} & 38.5 & 32.74\tabularnewline
    \hline 
    \textbf{Our model} & \textbf{62.8} & 79.7 & \textbf{70.1} & \textbf{32.8} & \textbf{55.6}  & \textbf{39.6} & 28.1 & \textbf{48.1} &\textbf{33.9}  \tabularnewline
    \hline 
    \end{tabular}}
    \label{tab:comparison_results}
\end{table*}

\subsubsection{Comparison with existing non-hybrid approaches:} To isolate the contribution of our hybrid extractive-abstractive summarization approach, we conduct a comparative evaluation against non-hybrid MDS models. As presented in Table~\ref{tab:comparison_results_without_hybrid}, our proposed hybrid model are superior to the baselines across all ROUGE metrics, particularly F1-score ROUGE-1 (70.1\% vs. KL's 60.2\%) and Recall (ROUGE-1 and ROUGE-2). While our ROUGE-2 F1-score is slightly lower than MART's (39.6\% vs. 41.6\%), the difference is not significant. This demonstrates the effectiveness of our approach in capturing the salient information from Vietnamese text documents and generating high-quality summaries.

\begin{table*}[ht]
    \centering
    \caption{Comparative results of models on VN-MDS dataset.}
    % Preview source code for paragraph 0

    % Preview source code for paragraph 0

\begin{tabular}{|c|c|c|c|c|}
\hline 
\multirow{2}{*}{Model} & \multicolumn{2}{c|}{ROUGE-1} & \multicolumn{2}{c|}{ROUGE-2}\tabularnewline
\cline{2-5} \cline{3-5} \cline{4-5} \cline{5-5} 
 & Recall & F1 score & Recall & F1 score\tabularnewline
\hline 
MART~\cite{nguyen2018towards} & 70.2\% & 49.8\% & 49.6\% & \textbf{41.6\%} \tabularnewline
\hline 
KL~\cite{nguyen2018towards} & 65.1\% & 60.2\% & 38.0\% & 40.4\%\tabularnewline
\hline 
LSA~\cite{nguyen2018towards} & 62.5\% & 49.2\% & 36.0\% & 39.2\%\tabularnewline
\hline 
\textbf{Our model} & \textbf{79.7\%} & \textbf{70.1\%} & \textbf{55.6\%} & 39.6\% \tabularnewline
\hline 
\end{tabular}
    \label{tab:comparison_results_without_hybrid}
\end{table*}
\section{Conclusion}

We have proposed a new Vietnamese multi-document summarization model combining extractive and abstractive techniques in a pipeline architecture. For extraction, we apply SBERT to identify salient sentences. These extracted sentences are then input to the VBD-LLaMA2-7B-50b language model for abstractive summarization. Experiments on the VN-MDS dataset demonstrate the efficacy of our approach, achieving competitive results over the existing methods.

Moving forward, we aim to evaluate our model on additional Vietnamese datasets. We also plan to explore alternative deep learning models to enhance extraction and abstractive generation. In addition, we will investigate the application of our model to unstructured data. This is a challenging task due to the lack of a Vietnamese dataset for evaluating models on unstructured data.
\bibliographystyle{plainnat}
\bibliography{refs}

\begin{thebibliography}{22}
\providecommand{\natexlab}[1]{#1}
\providecommand{\url}[1]{\texttt{#1}}
\expandafter\ifx\csname urlstyle\endcsname\relax
  \providecommand{\doi}[1]{doi: #1}\else
  \providecommand{\doi}{doi: \begingroup \urlstyle{rm}\Url}\fi

\bibitem[Devlin et~al.(2018)Devlin, Chang, Lee, and Toutanova]{devlin2018bert}
Jacob Devlin, Ming-Wei Chang, Kenton Lee, and Kristina Toutanova.
\newblock Bert: Pre-training of deep bidirectional transformers for language understanding.
\newblock \emph{arXiv preprint arXiv:1810.04805}, 2018.

\bibitem[Erkan and Radev(2004)]{erkan2004lexrank}
G{\"u}nes Erkan and Dragomir~R Radev.
\newblock Lexrank: Graph-based lexical centrality as salience in text summarization.
\newblock \emph{Journal of artificial intelligence research}, 22:\penalty0 457--479, 2004.

\bibitem[Gao et~al.(2021)Gao, Liu, Li, Guo, and Xiao]{gao2021extractive}
Yan Gao, Zhengtao Liu, Juan Li, Fan Guo, and Fei Xiao.
\newblock Extractive-abstractive summarization of judgment documents using multiple attention networks.
\newblock In \emph{Logic and Argumentation: 4th International Conference, CLAR 2021, Hangzhou, China, October 20--22, 2021, Proceedings 4}, pages 486--494. Springer, 2021.

\bibitem[Jin et~al.(2020)Jin, Wang, and Wan]{jin2020multi}
Hanqi Jin, Tianming Wang, and Xiaojun Wan.
\newblock Multi-granularity interaction network for extractive and abstractive multi-document summarization.
\newblock In \emph{Proceedings of the 58th annual meeting of the association for computational linguistics}, pages 6244--6254, 2020.

\bibitem[Liu et~al.(2021)Liu, Gao, Li, and Yang]{liu2021combined}
Wenfeng Liu, Yaling Gao, Jinming Li, and Yuzhen Yang.
\newblock A combined extractive with abstractive model for summarization.
\newblock \emph{IEEE Access}, 9:\penalty0 43970--43980, 2021.

\bibitem[Liu et~al.(2019)Liu, Ott, Goyal, Du, Joshi, Chen, Levy, Lewis, Zettlemoyer, and Stoyanov]{liu2019roberta}
Yinhan Liu, Myle Ott, Naman Goyal, Jingfei Du, Mandar Joshi, Danqi Chen, Omer Levy, Mike Lewis, Luke Zettlemoyer, and Veselin Stoyanov.
\newblock Roberta: A robustly optimized bert pretraining approach.
\newblock \emph{arXiv preprint arXiv:1907.11692}, 2019.

\bibitem[Manh et~al.(2019)Manh, Le~Thanh, and Minh]{manh2019extractive}
Hai~Cao Manh, Huong Le~Thanh, and Tuan~Luu Minh.
\newblock Extractive multi-document summarization using k-means, centroid-based method, mmr, and sentence position.
\newblock In \emph{Proceedings of the 10th International Symposium on Information and Communication Technology}, pages 29--35, 2019.

\bibitem[Mihalcea and Tarau(2004)]{mihalcea2004textrank}
Rada Mihalcea and Paul Tarau.
\newblock Textrank: Bringing order into text.
\newblock In \emph{Proceedings of the 2004 conference on empirical methods in natural language processing}, pages 404--411, 2004.

\bibitem[Nallapati et~al.(2016)Nallapati, Xiang, and Zhou]{nallapati2016sequence}
Ramesh Nallapati, Bing Xiang, and Bowen Zhou.
\newblock Sequence-to-sequence rnns for text summarization.
\newblock 2016.

\bibitem[Narayan et~al.(2018)Narayan, Cohen, and Lapata]{narayan2018ranking}
Shashi Narayan, Shay~B Cohen, and Mirella Lapata.
\newblock Ranking sentences for extractive summarization with reinforcement learning.
\newblock \emph{arXiv preprint arXiv:1802.08636}, 2018.

\bibitem[Nguyen et~al.(2018)Nguyen, Nguyen, Nguyen, et~al.]{nguyen2018towards}
Minh-Tien Nguyen, Hoang-Diep Nguyen, Van-Hau Nguyen, et~al.
\newblock Towards state-of-the-art baselines for vietnamese multi-document summarization.
\newblock In \emph{2018 10th International Conference on Knowledge and Systems Engineering (KSE)}, pages 85--90. IEEE, 2018.

\bibitem[QuangPH et~al.(2024)QuangPH, KietBS, and MinhTT]{VBDLLaMA2Chat2024}
QuangPH, KietBS, and MinhTT.
\newblock Vbd-llama2-chat - a conversationally-tuned llama2 for vietnamese.
\newblock \url{https://huggingface.co/LR-AI-Labs/vbd-llama2-7B-50b-chat}, 2024.
\newblock VinBigData Research.

\bibitem[Reimers and Gurevych(2019)]{reimers2019sentence}
Nils Reimers and Iryna Gurevych.
\newblock Sentence-bert: Sentence embeddings using siamese bert-networks.
\newblock \emph{arXiv preprint arXiv:1908.10084}, 2019.

\bibitem[Savery et~al.(2020)Savery, Abacha, Gayen, and Demner-Fushman]{savery2020question}
Max Savery, Asma~Ben Abacha, Soumya Gayen, and Dina Demner-Fushman.
\newblock Question-driven summarization of answers to consumer health questions.
\newblock \emph{Scientific Data}, 7\penalty0 (1):\penalty0 322, 2020.

\bibitem[Song et~al.(2019)Song, Huang, and Ruan]{song2019abstractive}
Shengli Song, Haitao Huang, and Tongxiao Ruan.
\newblock Abstractive text summarization using lstm-cnn based deep learning.
\newblock \emph{Multimedia Tools and Applications}, 78:\penalty0 857--875, 2019.

\bibitem[Thanh et~al.(2022)Thanh, Ngo, Ngo, Tran, and Ha]{thanh2022graph}
Tam-Doan Thanh, Xuan-Bach Ngo, Doan-Thinh Ngo, Mai-Vu Tran, and Quang-Thuy Ha.
\newblock A graph and phobert based vietnamese extractive and abstractive multi-document summarization frame.
\newblock In \emph{2022 RIVF International Conference on Computing and Communication Technologies (RIVF)}, pages 482--487. IEEE, 2022.

\bibitem[To et~al.(2021)To, Van~Nguyen, Nguyen, and Nguyen]{to2021monolingual}
Huy~Quoc To, Kiet Van~Nguyen, Ngan Luu-Thuy Nguyen, and Anh Gia-Tuan Nguyen.
\newblock Monolingual vs multilingual bertology for vietnamese extractive multi-document summarization.
\newblock In \emph{Proceedings of the 35th Pacific Asia Conference on Language, Information and Computation}, pages 692--699, 2021.

\bibitem[Touvron et~al.(2023{\natexlab{a}})Touvron, Lavril, Izacard, Martinet, Lachaux, Lacroix, Rozi{\`e}re, Goyal, Hambro, Azhar, et~al.]{touvron2023llama}
Hugo Touvron, Thibaut Lavril, Gautier Izacard, Xavier Martinet, Marie-Anne Lachaux, Timoth{\'e}e Lacroix, Baptiste Rozi{\`e}re, Naman Goyal, Eric Hambro, Faisal Azhar, et~al.
\newblock Llama: Open and efficient foundation language models.
\newblock \emph{arXiv preprint arXiv:2302.13971}, 2023{\natexlab{a}}.

\bibitem[Touvron et~al.(2023{\natexlab{b}})Touvron, Lavril, Izacard, Martinet, Lachaux, Lacroix, Rozi{\`e}re, Goyal, Hambro, Azhar, et~al.]{touvron2023llama1}
Hugo Touvron, Thibaut Lavril, Gautier Izacard, Xavier Martinet, Marie-Anne Lachaux, Timoth{\'e}e Lacroix, Baptiste Rozi{\`e}re, Naman Goyal, Eric Hambro, Faisal Azhar, et~al.
\newblock Llama: Open and efficient foundation language models.
\newblock \emph{arXiv preprint arXiv:2302.13971}, 2023{\natexlab{b}}.

\bibitem[Touvron et~al.(2023{\natexlab{c}})Touvron, Martin, Stone, Albert, Almahairi, Babaei, Bashlykov, Batra, Bhargava, Bhosale, et~al.]{touvron2023llama2}
Hugo Touvron, Louis Martin, Kevin Stone, Peter Albert, Amjad Almahairi, Yasmine Babaei, Nikolay Bashlykov, Soumya Batra, Prajjwal Bhargava, Shruti Bhosale, et~al.
\newblock Llama 2: Open foundation and fine-tuned chat models.
\newblock \emph{arXiv preprint arXiv:2307.09288}, 2023{\natexlab{c}}.

\bibitem[Tretyak and Stepanov(2020)]{tretyak2020combination}
Vladislav Tretyak and Denis Stepanov.
\newblock Combination of abstractive and extractive approaches for summarization of long scientific texts.
\newblock \emph{arXiv preprint arXiv:2006.05354}, 2020.

\bibitem[Verma and Nidhi(2017)]{verma2017extractive}
Sukriti Verma and Vagisha Nidhi.
\newblock Extractive summarization using deep learning.
\newblock \emph{arXiv preprint arXiv:1708.04439}, 2017.

\end{thebibliography}

% \printbibliography
% \bibliography{refs}
% %
% \begin{thebibliography}{8}
% \bibitem{ref_article1}
% Author, F.: Article title. Journal \textbf{2}(5), 99--110 (2016)

% \bibitem{ref_lncs1}
% Author, F., Author, S.: Title of a proceedings paper. In: Editor,
% F., Editor, S. (eds.) CONFERENCE 2016, LNCS, vol. 9999, pp. 1--13.
% Springer, Heidelberg (2016). \doi{10.10007/1234567890}

% \bibitem{ref_book1}
% Author, F., Author, S., Author, T.: Book title. 2nd edn. Publisher,
% Location (1999)

% \bibitem{ref_proc1}
% Author, A.-B.: Contribution title. In: 9th International Proceedings
% on Proceedings, pp. 1--2. Publisher, Location (2010)

% \bibitem{ref_url1}
% LNCS Homepage, \url{http://www.springer.com/lncs}. Last accessed 4
% Oct 2017
% \end{thebibliography}

\end{document}